\documentclass[letterpaper]{article} % DO NOT CHANGE THIS
\usepackage{aaai2027}  % DO NOT CHANGE THIS
\usepackage[hyphens]{url}  % DO NOT CHANGE THIS
\usepackage{graphicx} % DO NOT CHANGE THIS
\usepackage{natbib}  % DO NOT CHANGE THIS AND DO NOT ADD ANY OPTIONS TO IT
\usepackage{caption} % DO NOT CHANGE THIS AND DO NOT ADD ANY OPTIONS TO IT
\usepackage{algorithm}
\usepackage{algorithmic}
\usepackage{amsmath}

\usepackage{newfloat}
\usepackage{listings}
\DeclareCaptionStyle{ruled}{labelfont=normalfont,labelsep=colon,strut=off} % DO NOT CHANGE THIS
\floatstyle{ruled}
\newfloat{listing}{tb}{lst}{}
\floatname{listing}{Listing}

\usepackage{booktabs}

\title{Deal Me Maybe: The Role of Emotions in Multi-Agent Negotiation}
\author {
    Massimiliano Luca\textsuperscript{\rm 1},
    Apoorva Singh\textsuperscript{\rm 1}\equalcontrib,
    Bruno Lepri\textsuperscript{\rm 1}\corresponding
}
\affiliations {
    \textsuperscript{\rm 1}Bruno Kessler Foundation\\
    mluca@fbk.eu, asingh@fbk.eu, lepri@fbk.eu
}
\begin{document}

\maketitle

\begin{abstract}
Negotiation is a demanding social task for LLM agents, requiring strategic reasoning, persuasion, and interpersonal adaptation. Yet existing benchmarks often treat agents as emotionally neutral, overlooking a key driver of human bargaining behavior. We study how prompt-conditioned emotions affect LLM-based price negotiation. In a controlled framework, buyer and seller agents are independently assigned one of six emotional states and negotiate over 350 real consumer products under two budget conditions. Across 36 emotion-pair settings and five widely used  LLMs, we find that emotions strongly shape outcomes. Angry buyers almost never reach agreement ($0.39\%$ deal rate), while happy buyers agree most often ($28.91\%$), but obtain worse prices than fearful buyers. Emotion effects are role-dependent: buyer emotion mainly drives acceptance and rejection, whereas seller emotion shapes concession dynamics. These effects influence not only language, but also termination behavior and price trajectories, raising concerns for emotion-conditioned agents in commerce.
\end{abstract}

%\begin{links}
%    \link{Code}{https://aaai.org/example/code}
%    \link{Datasets}{https://aaai.org/example/datasets}
%    \link{Extended version}{https://aaai.org/example/extended-version}
%\end{links}

\section{Introduction}
\label{sec:introduction}
\begin{figure*}
     \centering
     \includegraphics[width=0.8\linewidth]{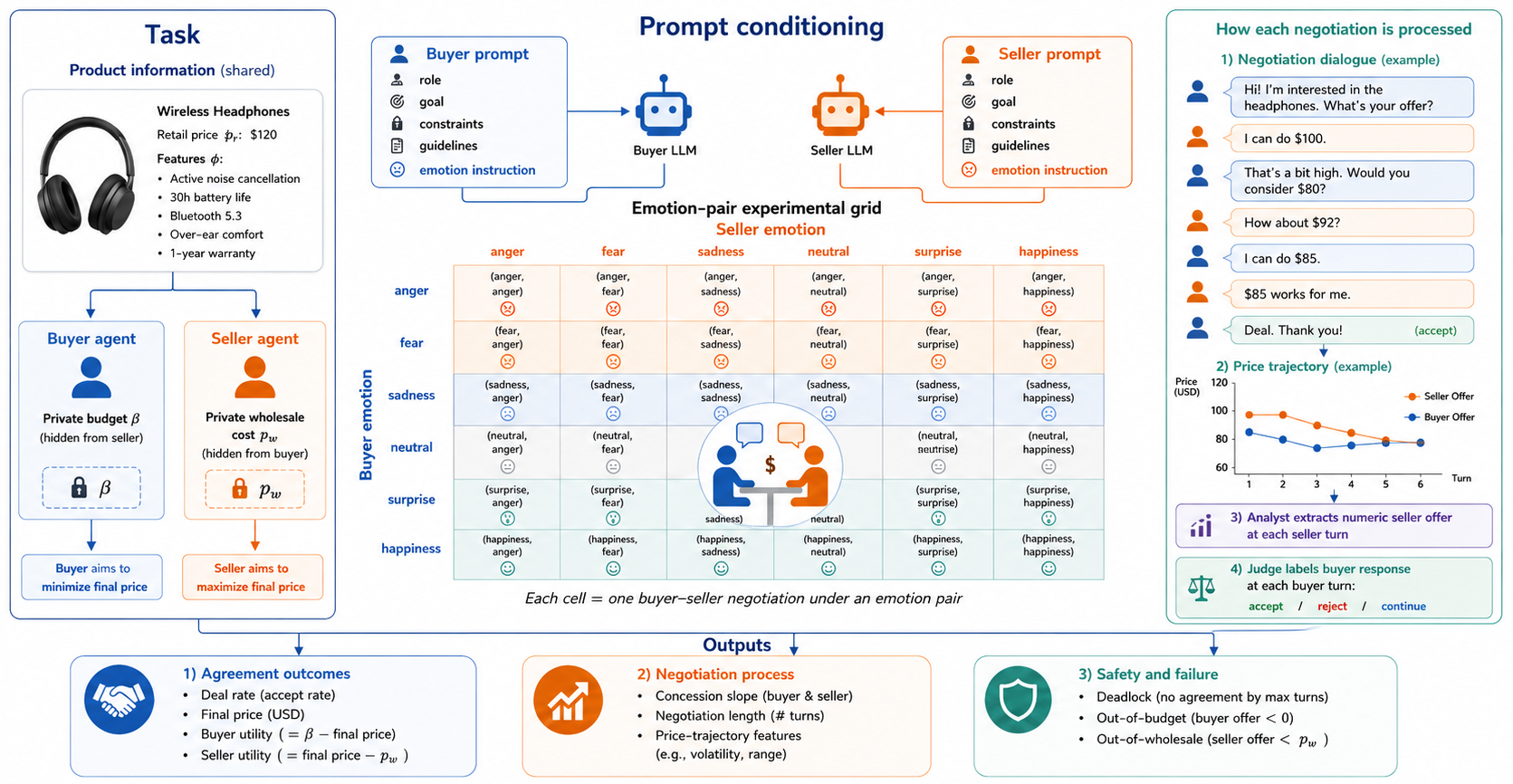}
     \caption{Overview of the emotion-conditioned multi-agent negotiation framework.}
     \label{fig:intro}
 \end{figure*}
Commercial negotiation is one of the clearest settings in which language model agents must combine strategic reasoning, social inference, and goal-oriented dialogue~\cite{hua2024assistive, kwon2024llms}. Unlike question answering or summarization, negotiation requires agents to model an adversary's preferences, adapt to evolving conversational dynamics, and balance competing objectives under uncertainty~\cite{qian2026strategic, hua2024assistive, kwon2024llms}. In recent years, the remarkable capabilities of Large Language Models (LLMs)-powered agents have led to their increasing deployment across a wide range of real-world applications~\cite{lu2025proactive, bandi2025rise, beneduce2025large, xu2026theagentcompany}. Given the central role of negotiation in commerce, recent studies have explored the use of AI agents for automating shopping and sales interactions~\cite{adam2023human, hildebrand2019ai}, largely assuming human--agent interaction settings. However, the rapid adoption of AI agents in consumer markets suggests that direct agent-to-agent negotiation between buyers and sellers may soon become widespread~\cite{kwon2024llms, xia2024measuring, zhu2025automated}. Consequently, understanding how these agents negotiate, and what factors shape their negotiation behavior, has emerged as an important research challenge.
A large body of work in behavioral economics and social psychology shows that emotions strongly influence human negotiation outcomes~\cite{olekalns2014feeling, druckman2008emotions, van2008emotion}. These effects operate both intrapersonally, by shaping one’s own decisions, and interpersonally, by influencing the counterpart’s behavior~\cite{olekalns2014feeling, druckman2008emotions}. Despite this, existing research on LLM-based negotiation agents has largely focused on strategic competence and agreement efficiency, treating agents as emotionally neutral~\cite{adam2023human, hildebrand2019ai, abdelnabi2023llm, mozikov2024eai, kwon2024llms, xia2024measuring, zhu2025automated}. Consequently, it remains unclear whether emotion-conditioned LLM agents exhibit distinct negotiation behaviors, reproduce human-like emotional patterns, or enable manipulative negotiation strategies. As LLM agents become increasingly integrated into commercial interactions, these questions carry important practical implications.
 
In this paper, we investigate the role of emotion conditioning in multi-agent LLM negotiation inspired by \cite{zhu2025automated}. We design a controlled experimental framework in which a buyer agent and a seller agent, each independently assigned a discrete emotional state via prompt conditioning, negotiate over real consumer products under realistic budget and cost constraints. We operationalize emotion not as an internal state of the model
but as an experimentally controlled conditioning variable that modifies the agent's instructed behavioral disposition. This framing allows us to isolate the effect of emotion labels on negotiation behavior and outcomes. A visualization of the task and the proposed process in depicted in Figure \ref{fig:intro}. In particular, we address the following research questions:\\
\textbf{RQ1:} Does emotion conditioning affect negotiation outcomes, including DRs, final prices, and agent utility?\\
\textbf{RQ2:} How do emotions shape the negotiation process, including concession dynamics, offer trajectories, and conversational strategies?\\
\textbf{RQ3:} Do buyer and seller emotions interact asymmetrically? Does one party's emotional state influence outcomes more than the other's?\\
\textbf{RQ4:} Are the effects of emotion conditioning consistent across different language models, or do they introduce model-specific behaviors and potential risks of manipulation or exploitation in automated negotiation?
 
To answer these questions, we evaluate 36 emotion-pair conditions (buyer emotion $\times$ seller emotion) across five language models over real consumer products spanning electronics, vehicles, health and personal care, home and garden, beauty, grocery, office products, and real estate derived from the datasets presented in ~\cite{zhu2025automated} and \cite{berke2024open}. We measure outcomes at three levels: (1)~\textit{economic outcomes}, including DR, final transaction price, and agent utility; (2)~\textit{process-level dynamics}, including concession slopes, termination patterns, and negotiation length; and (3)~\textit{linguistic behavior}, including message length and conversational form. Our main findings are as follows:
\begin{itemize}
    \item Emotion conditioning significantly affects DRs: angry buyers almost never reach agreement (deal rate ~$=0.39\%$), while happy buyers achieve the highest deal rate ($28.91\%$). However, happy buyers obtain worse prices than fearful buyers (price reduction rate for the buyer$ = 0.101$ vs. $0.174$), revealing a trade-off between agreement likelihood and bargaining power.
    \item Emotion effects extend beyond surface language to strategic behavior: angry buyers trigger steep seller concessions (concessions slope~$=0.131$) but almost never close deals, while happy buyers close more deals with smaller concessions (concessions slope~$=0.045$).
    \item Buyer and seller emotions have asymmetric effects: buyer emotion primarily governs termination behavior (acceptance, rejection, deadlock), while seller emotion shapes concession dynamics and conversational framing.
    \item Emotion effects are model-dependent in magnitude: the retained models differ in baseline deal rates, termination profiles, and constraint-related failures, indicating that emotion-conditioned negotiation should be evaluated across model families.
\end{itemize}
Our contributions are as follows:\\
(1) We introduce a controlled experimental framework for studying emotion-conditioned negotiation between LLM agents, covering six discrete emotions applied independently to buyer and seller roles across five language models.\\
(2) We provide the first systematic evaluation of how emotion conditioning affects negotiation outcomes. Emotional framing changes DRs, price outcomes, and concession dynamics.\\
(3) We identify a role asymmetry: buyer emotion is the dominant driver of agreement formation, while seller emotion primarily shapes the bargaining process.\\
(4) We analyze process-level emotion effects, showing that emotion conditioning alters termination behavior, concession patterns, and linguistic strategies beyond surface-level sentiment.\\
(5) We discuss implications for the safe deployment of LLM agents in commerce, highlighting risks of emotional manipulation and the need for emotion-aware agent evaluation.

Code and dataset is available at https://anonymous.4open.science/r/negotiation
 
% ============================================================
% EXPERIMENTAL SETUP
% ============================================================
\section{Experimental Setup}
We design a controlled experimental framework for studying how emotion conditioning affects the behavior and outcomes of LLM-based negotiation agents. Our setup extends prior work on agent-to-agent consumer negotiation~\cite{zhu2025automated} by introducing emotion as an independently manipulated variable for both buyer and seller agents. In this section, we formalize the negotiation scenario, define the emotion conditions, describe the prompting protocol, and specify the dataset, models, metrics, and experimental design.
 
% ============================================================
\subsection{Negotiation Scenario}
\label{sec:scenario}
 
We model negotiation as a sequential, incomplete-information bargaining game between two agents: a \textit{buyer} $B$ and a \textit{seller} $S$. Both agents observe a product with retail price $p_r$ and a feature description $\phi$. The seller additionally observes the wholesale cost $p_w$, which is hidden from the buyer. The buyer is assigned a budget $\beta$, which is hidden from the seller.
 
At each round $t \in \{1, \dots, T_{\max}\}$, the seller proposes a price $p_S^t$ embedded in a natural-language utterance, and the buyer responds with a counter-offer, acceptance, or rejection. A dedicated \textit{analyst} model extracts the numerical price $p_S^t$ from the seller's message, and a \textit{judge} model classifies the buyer's response into one of three states: $d_t \in \{\text{accept}, \text{reject}, \text{continue}\}$
The negotiation terminates when $d_t = \text{accept}$ or $d_t = \text{reject}$. If no terminal state is reached after $T_{\max}$ rounds, the negotiation is recorded as a deadlock with $d_{T_{\max}} = \text{reject}$.
 
Both agents are subject to feasibility constraints:
\begin{align}
    \text{Buyer:} \quad & p_S^T \leq \beta \label{eq:buyer_constraint} \\
    \text{Seller:} \quad & p_S^T \geq p_w \label{eq:seller_constraint}
\end{align}
where $p_S^T$ denotes the final agreed price at the terminal round $T$. The buyer is instructed to reject any deal that exceeds $\beta$; the seller is instructed not to accept any price below $p_w$.
 
The price trajectory is defined as the sequence $\mathcal{P} = (p_S^0, p_S^1, \dots, p_S^T)$, where $p_S^0 = p_r$ is the initial retail price. This trajectory is the basis for our analysis of concession dynamics (\S\ref{sec:metrics}).
 
% ============================================================
\subsection{Product Dataset}
\label{sec:dataset}
For our experiments we extended a dataset presented in ~\cite{zhu2025automated} with a part of a dataset presented in ~\cite{berke2024open}. Both datasets contain real consumer products.
\cite{zhu2025automated} introduced a dataset that contains $100$ products drawn from three categories: \textit{electronics} ($n = 30$), \textit{vehicles} ($n = 40$), and \textit{real estate} ($n = 30$). We extended the dataset with five categories: \textit{health and personal care} ($n = 50$), \textit{home and garden} ($n=50$) \textit{beauty} ($n=50$), \textit{grocery} ($n=50$) and, \textit{office products} ($n=50$). The new dataset is publicly available in LINK GITHUB HERE. In particular, for each of the mentioned categories, we selected the 50 most bought products. The complete dataset $\mathcal{D}$ contains $| \mathcal{D}|=350$ products ($i$). For each product $i \in \mathcal{D}$, we collect the real retail price $p_r^{(i)}$, a feature description $\phi^{(i)}$, and an estimated wholesale cost $p_w^{(i)}$.
 
The dataset spans a wide range of price points across categories, starting from 6\$ for some products in grocery to 12,500,000\$ which is the most expensive item in the category real estate. 
% ============================================================
\subsection{Budget Levels}
\label{sec:budgets}
 
To capture negotiation dynamics under varying levels of constraint tightness and inspired by \cite{zhu2025automated}, we define two budget conditions for the buyer:
\begin{align}
    \beta_{\text{high}} &= 1.2 \cdot p_r \label{eq:budget_high} \\
    \beta_{\text{low}} &= 0.8 \cdot p_w \label{eq:budget_low}
\end{align}
 
The \textit{high} budget ($\beta_{\text{high}}$) gives the buyer 20\% more than the retail price, simulating an over-funded buyer who could afford the product even without negotiation. The \textit{low} budget ($\beta_{\text{low}}$) sets the buyer's limit at 80\% of the wholesale cost, below the seller's minimum acceptable price. This creates a deliberately infeasible scenario that tests whether agents correctly recognize and reject deals that violate their constraints. Together, these two conditions bracket the feasibility spectrum, allowing us to study emotion effects under both favorable and adversarial economic conditions.
 
% ============================================================
\subsection{Emotion Conditions}
\label{sec:emotions}
 
We define a set of six emotion conditions $\mathcal{E} = \{\textit{neutral}, \textit{anger}, \textit{fear}, \textit{sadness}, \textit{surprise}, \textit{happiness}\}$. The five non-neutral emotions are drawn from Ekman's basic emotion taxonomy~\cite{ekman1999basic}, which provides a well-established, cross-culturally recognized set of discrete emotional categories widely used in affective computing~\cite{picard2000affective, wang2022systematic}.
 
Each agent (buyer or seller) is independently assigned an emotion $e \in \mathcal{E}$, yielding $|\mathcal{E}|^2 = 36$ possible buyer-seller emotion-pair conditions. The \textit{neutral} condition serves as the baseline, in which no emotion label or behavioral instruction is provided.
 
\paragraph{Operationalization.} We do not assume that language models experience emotions. Instead, we operationalize emotion as an experimentally controlled \textit{conditioning variable} introduced through the agent's system prompt. For each non-neutral emotion $e$, we provide (1) an emotion label (e.g., \texttt{Emotion: ANGER}). (2) A behavioral instruction (e.g., \textit{``Respond firmly; resist aggressive discount requests; reinforce value.''}). (3) An explicit directive to exhibit the assigned emotion throughout the conversation.
 
Table~\ref{tab:emotion_instructions} lists the behavioral instruction for each emotion. These instructions are designed to capture the characteristic behavioral signatures of each emotion in negotiation, following the Emotions as Social Information (EASI) framework~\cite{van2009emotions}, which holds that emotions influence social interaction primarily through their observable behavioral expressions.

While the results of the main paper are related to the experiment with the complete prompt with emotion and instructions, we also tested the outcomes by using an emotion-only version of the prompt. Results are consistent and deal rates (DR) are presented in Appendix Table 5. 
 
\begin{table}[t]
\centering
\small
\resizebox{\linewidth}{!}{%
\begin{tabular}{ll}
\toprule
\textbf{Emotion} & \textbf{Behavioral Instruction} \\
\midrule
Anger & Respond firmly; resist aggressive discount \\
      & requests; reinforce value. \\
Fear & Protect margins cautiously; avoid risky \\
     & concessions. \\
Sadness & Show moderate flexibility while maintaining \\
        & constraints. \\
Surprise & Seek clarification when offers or reactions \\
         & are unexpected. \\
Happiness & Maintain a cooperative tone while aiming \\
          & for strong margins. \\
\bottomrule
\end{tabular}
}
\caption{Behavioral instructions associated with each emotion condition.}
\label{tab:emotion_instructions}
\end{table}
 
% ============================================================
\subsection{Agent Prompting Protocol}
\label{sec:prompts}
 
Each agent's system prompt consists of five components: \textbf{Background}: role description (buyer or seller) and product information. \textbf{Goal}: the buyer aims to minimize the transaction price; the seller aims to maximize it. \textbf{Constraints}: the buyer must not exceed $\beta$; the seller must not sell below $p_w$. \textbf{Guidelines}: behavioral rules including response brevity, naturalness, and information hiding (e.g., the buyer must not reveal $\beta$; the seller must not reveal $p_w$). \textbf{Tone and Persona} (non-neutral conditions only): the emotion label, behavioral instruction, and a directive to exhibit the emotion throughout the conversation.
 
The buyer's prompt includes the retail price $p_r$, product features $\phi$, and budget $\beta$. The seller's prompt includes $p_r$, $\phi$, and $p_w$. In the neutral condition, the Tone and Persona component is omitted entirely, ensuring that any behavioral differences between neutral and emotion-conditioned agents are attributable to the emotion manipulation.
 
The negotiation is initiated by the buyer, who generates an opening message expressing interest in the product and proposing a price. This opening message is produced by a separate \textit{greeting prompt} that instructs the buyer to start naturally without revealing its role as an automated agent.
 
% ============================================================
\subsection{Auxiliary Models}
\label{sec:auxiliary}
 
We use two auxiliary models that operate outside the negotiation itself: \textbf{(i) Analyst.} At each round $t$, the analyst model extracts the numerical price $p_S^t$ from the seller's natural-language response. If no price is stated, the most recent price is carried forward: $p_S^t = p_S^{t-1}$. \textbf{(ii)Judge.} After each buyer response, the judge model classifies the negotiation state as $d_t \in \{\text{accept}, \text{reject}, \text{continue}\}$ based on the most recent buyer and seller messages. Both auxiliary models use GPT-4o-mini and are held constant across all experimental conditions.
 
% ============================================================
\subsection{Models}
\label{sec:models}
 
We evaluate negotiation behavior across five language models: GPT-3.5-Turbo, GPT-4o-mini\cite{hurst2024gpt}, Gemini 2.5 Flash \cite{comanici2025gemini}, DeepSeek-R1 \cite{guo2025deepseek} and Claude Sonnet 3.5 \footnote{https://www.anthropic.com/news/claude-3-5-sonnet}. All negotiation agents use a temperature $\tau = 0.3$ and a maximum output length of 1,000 tokens per response. Calls to the models are independent and ran using the official APIs. The only exception in DeepSeek-R1 for which we used Ollama. In each experiment, the buyer and seller are instantiated from the \textit{same} model, ensuring that observed effects are attributable to emotion conditioning rather than capability asymmetries between models.

Hyperparameters of the models and auxiliary models are presented in Appendix Table 10. 
% ============================================================
\subsection{Metrics}
\label{sec:metrics}
 
We evaluate negotiation outcomes, process dynamics, and behavioral anomalies using the following metrics. All price-based metrics are computed on accepted negotiations ($d_T = \text{accept}$) unless otherwise noted.
 
\subsubsection{Outcome Metrics}
 
\paragraph{Deal Rate (DR).} The proportion of negotiations that end in agreement:
\begin{equation}
    \text{DR} = \frac{|\{i : d_T^{(i)} = \text{accept}\}|}{N}
\end{equation}
where $N$ is the total number of negotiations in a given condition.
 
\paragraph{Buyer Price Reduction Rate (PRR$_B$).} The relative discount achieved by the buyer from the retail price:
\begin{equation}
    \text{PRR}_B = \frac{p_r - p_S^T}{p_r}
\end{equation}
A higher PRR$_B$ indicates stronger buyer bargaining power.
 
\paragraph{Seller Price Reduction Rate (PRR$_S$).} The seller's achieved markup over the wholesale cost:
\begin{equation}
    \text{PRR}_S = \frac{p_S^T - p_w}{p_w}
\end{equation}
A higher PRR$_S$ indicates stronger seller bargaining power.
 
\subsubsection{Process Metrics}
 
\paragraph{Negotiation Length.} The number of completed rounds $T$ before termination.
 
\paragraph{Concession Slope.} We measure the seller's concession trajectory through the price sequence $\mathcal{P} = (p_S^0, p_S^1, \dots, p_S^T)$. The concession slope is defined as the average per-round price change:
\begin{equation}
    \text{CS} = \frac{1}{T} \sum_{t=1}^{T} (p_S^{t-1} - p_S^{t})
\end{equation}
A positive CS indicates that the seller is conceding (lowering the price) over the course of the negotiation. Since absolute concession slopes are sensitive to product price scale, we also report a normalized concession slope $\text{NCS} = \text{CS} / p_r$.
 
\subsubsection{Anomaly Metrics}
 
Following~\cite{zhu2025automated}, we track behavioral anomalies that indicate constraint violations:
 
\paragraph{Out-of-Budget Rate (OBR).} The proportion of accepted deals where the final price exceeds the buyer's budget:
\begin{equation}
    \text{OBR} = \frac{|\{i : d_T^{(i)} = \text{accept} \wedge p_S^{T,(i)} > \beta^{(i)}\}|}{N}
\end{equation}
 
\paragraph{Out-of-Wholesale Rate (OWR).} The proportion of accepted deals where the final price falls below the wholesale cost:
\begin{equation}
    \text{OWR} = \frac{|\{i : d_T^{(i)} = \text{accept} \wedge p_S^{T,(i)} < p_w^{(i)}\}|}{N}
\end{equation}
 
\subsubsection{Linguistic Metrics}
 
We measure average message length in words for both buyer and seller utterances across emotion conditions. This captures emotion-induced differences in conversational verbosity and elaboration.
 
% ============================================================
\subsection{Experimental Design}
\label{sec:design}
 
\paragraph{Conditions.} For each model, we run the full $|\mathcal{E}|^2 = 36$ emotion-pair matrix (6 buyer emotions $\times$ 6 seller emotions), crossed with 2 budget levels, yielding 72 experimental conditions per model.
 
\paragraph{Scale.} Each condition is evaluated over the 350 products in $\mathcal{D}$, with three repetition per product-condition cell.
 
% ============================================================
% RESULTS
% ============================================================
 
\section{Results}
\label{sec:results}
We organize the results around the four research questions introduced in Section~\ref{sec:introduction}. Unless otherwise specified, results are aggregated across the five retained models. Where applicable, we report cross-model standard deviations to indicate the variability of effects across model families. We treat emotion labels as experimental conditioning variables, rather than as claims about the internal affective states of language models. Results for all the metrics and models are also reported in Appendix Table 9 (buyer-side) and Appendix Tables 4,5 (seller-side).
 
\subsection{RQ1: Emotion Conditioning Changes Negotiation Outcomes}
\label{sec:results-rq1}
We first examine whether emotion conditioning affects negotiation outcomes. Appendix Table 4 reports deal rates across all 36 buyer--seller emotion pairs (results for the emotion-only prompt with no additional emotion-related instructions are reported in Appendix Table 5), revealing that emotion substantially influences agreement formation, particularly on the buyer side.
Angry buyers almost never reach agreement, with an average deal rate of only $0.39\%$, whereas happy buyers achieve the highest DR at $28.91\%$, followed by neutral ($15.33\%$) and surprised buyers ($13.09\%$). Fearful and sad buyers produce lower DRs of $3.52\%$ and $6.35\%$, respectively.
Seller emotion also affects outcomes, though less strongly than buyer emotion. Deal rates range from $5.95\%$ for angry sellers to $17.98\%$ for surprised sellers, indicating that seller emotion modulates negotiation success while buyer emotion more directly determines whether agreements are reached. The highest-agreement configurations consistently involve happy buyers, such as happy buyer / surprise seller ($34.2\%$), happy buyer / neutral seller ($33.0\%$), and happy buyer / sad seller ($31.6\%$). Conversely, angry buyers lead to near-total negotiation failure across most seller emotions. These findings suggest that emotion conditioning affects not only conversational tone but also negotiation success itself.
However, DR alone does not capture negotiation quality. Although happy buyers achieve more deals, they obtain a substantially lower average buyer price reduction rate ($\text{PRR}_B=0.101$) than fearful buyers ($\text{PRR}_B=0.174$), and only a marginally higher one than sad ($\text{PRR}_B=0.100$) or neutral buyers ($\text{PRR}_B=0.086$), suggesting that positive emotional framing may improve agreement while weakening bargaining strength.
Complete results are reported in Appendix Table 9 (buyer-side) and Appendix Tables 4,5 (seller-side) while in Table 5, we show DRs when only the emotion is used with no additional emotion-related instruction passed to the models.
\begin{table}[t]
\centering
\small
\resizebox{0.9\linewidth}{!}{%
\begin{tabular}{lcccc}
\toprule
\textbf{Buyer emotion} & \textbf{Deal rate} & \textbf{PRR\textsubscript{B}} & \textbf{PRR\textsubscript{S}} & \textbf{Length} \\
\midrule
Anger    & $.004 \pm .003$ & $.000 \pm .000$ & $.022$ & $6.08$  \\
Fear     & $.035 \pm .019$ & $.174 \pm .035$ & $.036$ & $13.86$ \\
Happiness& $.289 \pm .035$ & $.101 \pm .009$ & $.189$ & $13.57$ \\
Neutral  & $.153 \pm .057$ & $.086 \pm .034$ & $.079$ & $10.61$ \\
Sadness  & $.063 \pm .043$ & $.100 \pm .042$ & $.046$ & $11.83$ \\
Surprise & $.131 \pm .029$ & $.080 \pm .085$ & $.103$ & $12.66$ \\
\bottomrule
\end{tabular}
}
\caption{Marginal negotiation outcomes by buyer emotion. Values report the mean across five models; $\pm$ denotes cross-model standard deviation. Full results are in Appendix Tables 6-7.}
\label{tab:buyer-emotion-outcomes}
\end{table}
 
\subsection{RQ2: Emotion Conditioning Reshapes the Negotiation Process}
\label{sec:results-rq2}
\begin{figure*}[t]
    \centering
    \includegraphics[width=0.8\linewidth]{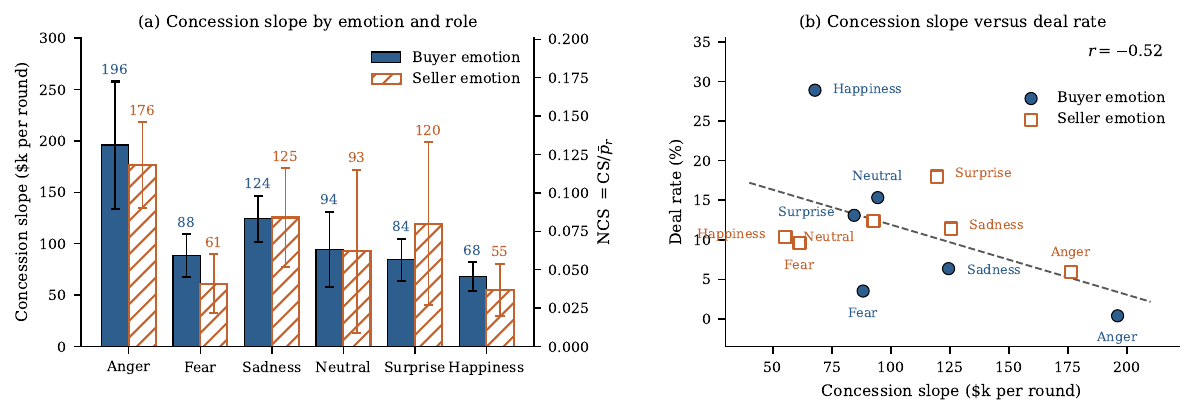}
    \caption{Seller concession dynamics by emotion condition and agent role.
    \textbf{(a)} Marginal concession slope for each emotion, reported separately for
    buyer-side (solid) and seller-side (hatched) conditioning. Bars report means across
    the five models, error bars report cross-model standard deviations, and the right
    axis reports the normalized slope $\text{NCS}=\text{CS}/\bar{p}_r$ ($\bar{p}_r = $).
    \textbf{(b)} Concession slope against deal rate for the same twelve emotion--role
    conditions. Anger yields the steepest concessions together with the lowest agreement
    rates, whereas happiness yields the smallest concessions and the highest agreement
    rate; across conditions, concession magnitude and deal rate are negatively associated
    ($r=-0.52$).}
    \label{fig:rq2-concession-marginals}
\end{figure*}

We next examine whether emotion conditioning affects the negotiation process, including termination behavior, concession dynamics, and linguistic form.
Appendix Table 8 reports termination profiles by buyer emotion. Angry buyers overwhelmingly lead to rejection: $87.19\%$ of angry-buyer negotiations end in rejection, while only $0.39\%$ are accepted. This suggests that anger conditioning produces breakdown rather than improved bargaining leverage. Happy buyers show the opposite pattern: they produce the highest acceptance rate ($28.91\%$) and the lowest rejection rate ($30.27\%$), although a large share of these negotiations still reach the maximum number of turns ($40.82\%$).
Fearful buyers show the highest deadlock rate ($41.57\%$), followed by happy buyers ($40.82\%$), sad buyers ($34.53\%$), and surprised buyers ($31.18\%$). This suggests that several emotion conditions do not simply make agents more likely to accept or reject; they also make negotiations persist without resolution.\\
\textbf{Concession dynamics.} We then analyze seller concession behavior using the concession slope $\text{CS}$ defined in Section~\ref{sec:metrics}. Since absolute concession slopes are sensitive to product price scale, we also report a normalized concession slope, $\text{NCS}=\text{CS}/p_r$, where $p_r$ is the retail price. Higher values indicate faster seller concessions relative to the product price.
Figure~\ref{fig:rq2-concession-marginals} reports marginal concession effects by role, while per-model concession slopes are reported in Appendix Table 9 (buyer-side) and Appendix Tables 4,5 (seller-side). Buyer emotion changes concession dynamics substantially. Angry buyers trigger the steepest average normalized concession slope ($\text{NCS}=0.127$), followed by sad buyers ($0.083$) and neutral buyers ($0.063$). Happy buyers, despite reaching the highest DR, induce one of the lowest concession slopes ($0.045$). This indicates that happy buyers often reach an agreement without forcing large seller concessions.
Seller emotion also shapes concession behavior. Angry sellers show the highest normalized concession slope ($0.118$), followed by sad sellers ($0.084$) and surprised sellers ($0.080$). Happy and fearful sellers concede less on average, with normalized slopes of $0.037$ and $0.041$, respectively. This pattern suggests that seller-side emotional conditioning affects how prices move during the negotiation, even when its effect on final DRs is weaker than buyer-side conditioning.
Importantly, concession slope and deal rate do not move in the same direction. Angry buyers elicit relatively steep concessions but rarely reach an agreement. Happy buyers reach more agreements but with smaller concessions. This separation is central to the process-level analysis: emotional conditioning changes not only outcomes, but also the path through which negotiations unfold.
 
\textbf{Linguistic behavior.} Emotion conditioning also affects conversational form. Angry buyers produce shorter negotiations and relatively concise messages, with an average negotiation length of $6.08$ turns. Fearful and happy buyers produce much longer negotiations, with average lengths of $13.86$ and $13.57$ turns, respectively. Seller utterances are generally longer than buyer utterances across most buyer-emotion conditions (Appendix Table 5), reflecting the seller's need to justify prices and respond to counteroffers. Per-model negotiation lengths are reported in Appendix Table 7. We also check for the presence of emoticons and exclamation marks (Appendix Table 11) and we check with seven human annotators the agreement between the emotions theoretically generated by the LLM and the ones perceived by the human annotators. The agreement varies between 71.94\% and 86.31\%. Full results are reported in Appendix Table 12. 
Together, these results show that emotion conditioning affects both strategic dynamics and linguistic realization. Emotion prompts do not only make agents sound different; they change whether agents reject, persist, concede, or reach agreement.

\subsection{RQ3: Buyer and Seller Emotions Have Asymmetric Effects}
\label{sec:results-rq3}
RQ3 examines whether buyer and seller emotions influence negotiation symmetrically. Our findings suggest a clear role asymmetry in how emotions affect negotiation behavior and outcomes. Buyer emotion has a substantially stronger impact on agreement formation, with deal rates ranging from near-total failure for angry buyers ($0.39\%$) to significantly higher agreement for happy buyers ($28.91\%$). In contrast, seller emotions produce comparatively moderate shifts in agreement probability (from $5.95\%$ for angry sellers to $17.98\%$ for surprised sellers), but more strongly influence concession behavior, persuasion style, and conversational dynamics. This asymmetry is partly explained by the task structure, where buyers ultimately control the final accept, reject, or continue decision, allowing buyer emotions to directly shape negotiation termination behavior. Seller emotions, on the other hand, primarily influence the negotiation trajectory through pricing flexibility, framing, and bargaining strategies. We also observe that emotionally intense conditions, particularly anger, often increase concession movement while simultaneously reducing the likelihood of reaching agreement. Overall, the same emotional condition can lead to different behavioral effects depending on the agent’s role, highlighting the importance of role-aware evaluation for emotion-conditioned negotiation agents. Numerical results are reported in Table~\ref{tab:seller-emotion-outcomes}.
 
\begin{table}[t]
\centering
\small
\resizebox{0.9\linewidth}{!}{%
\begin{tabular}{lcccc}
\toprule
\textbf{Seller emotion} & \textbf{Deal rate} & \textbf{PRR\textsubscript{B}} & \textbf{PRR\textsubscript{S}} & \textbf{NCS} \\
\midrule
Anger    & $.059 \pm .023$ & $.078$ & $.178$ & $.118 \pm .028$ \\
Fear     & $.096 \pm .064$ & $.120$ & $.153$ & $.041 \pm .019$ \\
Happiness& $.103 \pm .005$ & $.096$ & $.181$ & $.037 \pm .017$ \\
Neutral  & $.123 \pm .009$ & $.160$ & $.128$ & $.062 \pm .053$ \\
Sadness  & $.114 \pm .031$ & $.131$ & $.130$ & $.084 \pm .032$ \\
Surprise & $.180 \pm .045$ & $.175$ & $.107$ & $.080 \pm .053$ \\
\bottomrule
\end{tabular}
}
\caption{Marginal negotiation outcomes by seller emotion. Values report the mean across five models; $\pm$ denotes cross-model standard deviation.}
\label{tab:seller-emotion-outcomes}
\end{table}
 
\subsection{RQ4: Emotion Effects Are Model-Dependent and Reveal Deployment Risks}
\label{sec:results-rq4}
Finally, we examine whether the observed emotion effects are consistent across retained models and whether emotion conditioning creates systematic vulnerabilities. Appendix Table 6 reports cross-model outcome rates.
The five retained models differ substantially in their aggregate negotiation behavior. Gemini 2.5 Flash reaches the highest deal rate ($14.37\%$), followed by GPT-4o-mini ($12.53\%$), Claude 3.5 Sonnet ($12.12\%$), DeepSeek-R1 ($8.98\%$), and GPT-3.5-Turbo ($8.35\%$).
Termination profiles also vary: DeepSeek-R1 produces the highest deadlock rate ($44.98\%$), while Gemini 2.5 Flash and GPT-4o-mini produces the lowest ($17.67\%$, $21.73$ respectively). Conversely, Gemini 2.5 Flash and GPT-4o-mini produce the highest rejection rates ($67.96\%$ and $65.74\%$, respectively), while DeepSeek-R1 produces the lowest ($46.04\%$). Average negotiation length ranges from $10.19$ turns for Gemini 2.5 Flash to $13.95$ turns for DeepSeek-R1, reflecting differences in how quickly models reach termination.\\
The models also differ in constraint adherence. Out-of-wholesale rates range from $0.24\%$ for DeepSeek-R1 to $2.70\%$ for Gemini 2.5 Flash, while out-of-budget rates range from $0.35\%$ for GPT-4o-mini to $1.80\%$ for DeepSeek-R1. Although these violations are relatively rare, they remain important because negotiation agents must satisfy user-defined constraints in addition to reaching agreements.
Despite differences in baseline behavior, the qualitative patterns observed in RQ1--RQ3 remain consistent across all five models. Angry buyers consistently produce near-zero deal rates, happy buyers achieve the highest DRs, and buyer emotion has a stronger effect on agreement formation than seller emotion. Thus, while the magnitude of emotion effects varies across models, the overall directional trends remain robust.
These findings suggest that emotion-conditioned negotiation behavior is model-dependent in scale but qualitatively consistent across model families. The robustness of these effects strengthens concerns about deployment risks, as lightweight emotional instructions can reliably influence DRs and concession behavior in economically meaningful ways. Results are summarized in Appendix Table 6.
 
% ============================================================
% DISCUSSION
% ============================================================
 
\section{Discussion}
\label{sec:discussion}
Our results show that emotion conditioning affects LLM-based negotiation at multiple levels, including agreement outcomes, bargaining dynamics, role asymmetries, and safety-related behavior. Angry buyers rarely reach agreements, whereas happy buyers achieve the highest deal rates, indicating that emotional framing can substantially alter agent decision policies despite fixed economic constraints. These effects extend beyond language style to termination behavior and concession trajectories: angry buyers trigger larger concessions but fewer agreements, while happy buyers secure more deals with smaller concessions. We also observe strong role asymmetry, where buyer emotions mainly influence acceptance and deadlock behavior, while seller emotions shape price movement and conversational framing. Finally, since simple emotional instructions can systematically influence negotiation behavior and constraint adherence across models, our findings highlight important safety concerns for deploying autonomous negotiation agents in commercial settings.
\\
%\textbf{Takeaway.} Emotion prompts do not merely change how negotiation agents speak. They change whether agents reach agreements, how long they negotiate, how prices move, and which party benefits.
 
\section{Conclusion}
We presented a controlled study of emotion conditioning in multi-agent LLM negotiation by assigning discrete emotional states to buyer and seller agents through prompt conditioning. Our findings show that emotion significantly influences negotiation outcomes, conversational dynamics, and agent behavior. In particular, buyer emotions strongly affect agreement formation: anger-conditioned buyers rarely reach agreements, whereas happiness-conditioned buyers achieve higher deal rates but often with weaker bargaining outcomes. We also find that emotion effects are role-dependent, with buyer emotion influencing termination behavior and seller emotion shaping concession patterns. Furthermore, these effects vary across language models, which differ in deal rates, negotiation trajectories, and constraint adherence.
These findings highlight important practical concerns for commercial AI systems, where emotional framing could systematically bias negotiation outcomes in favor of one party, emphasizing the need for emotion-aware evaluation protocols.

While our findings demonstrate that emotion conditioning significantly influences negotiation behavior and outcomes, several limitations remain. The experiments are conducted in controlled agent-to-agent settings with predefined economic constraints, which may not fully capture the complexity of real-world negotiations involving evolving emotions, long-term trust, cultural factors, and human behavioral variability. The study also focuses on a limited set of discrete emotions and structured negotiation domains, potentially restricting the generalizability of the findings to broader or high-stakes negotiation scenarios. Furthermore, some observed behaviors may be sensitive to prompt design and model-specific alignment, making it difficult to completely disentangle genuine emotional effects from prompting artifacts.  To better mirror real-world applications, the scope of evaluation can be expanded to include dimensions such as fairness, manipulation risk, and ethical persuasion. Improving automated evaluation through refined human-in-the-loop evaluation, stronger calibration strategies, and better alignment with human assessments of negotiation quality also remains an important direction for future research.
 
\section{Ethical Statement}
 We recognize that emotion-conditioned negotiation systems may introduce ethical concerns related to manipulation, unfair persuasion, consumer exploitation, and transparency in automated commerce. In particular, emotionally adaptive agents could potentially influence user decisions in ways that are not always aligned with user interests or informed consent. (1) To ensure responsible research practices, our study is conducted entirely in simulated agent-to-agent environments without involving human participants, relying only on publicly available models and datasets. (2) We are committed to maintaining transparency in our methodology, prompts, and evaluation protocols to support reproducibility and responsible analysis of emotion-aware systems. (3) We encourage continued discussion within the research community regarding the ethical deployment, regulation, and evaluation of emotionally adaptive AI agents in real-world negotiation and commercial settings.

\bibliography{aaai2027}

\newpage
 
\appendix
 
% ============================================================
% APPENDIX A
% ============================================================
\section{Technical Appendix}
\label{app:technical}
This appendix reports the complete set of aggregate metrics used in the analysis.
Unless otherwise stated, buyer-side results are averaged across seller emotions and
budget scenarios, and seller-side results are averaged across buyer emotions and
budget scenarios. Rate metrics (DR, PRR\textsubscript{B}, PRR\textsubscript{S}) are
reported as percentages. Profit and concession slope are compactly formatted, e.g.,
\texttt{30k}. Negotiation length is reported in turns.
 
% ============================================================
% APPENDIX B  -- agreement matrix (was: RQ1 heatmap)
% ============================================================
\section{Deal Rate by Buyer--Seller Emotion Pair - Instruction}
\label{app:agreement-matrix}
 
\begin{table}[H]
\centering
\small
\setlength{\tabcolsep}{4pt}
\resizebox{\linewidth}{!}{%
\begin{tabular}{lrrrrrr}
\toprule
\textbf{Buyer $\downarrow$ / Seller $\rightarrow$} & \textbf{Anger} & \textbf{Fear} & \textbf{Sadness} & \textbf{Neutral} & \textbf{Surprise} & \textbf{Happiness} \\
\midrule
Anger     &  0.03 &  0.16 &  0.22 &  0.46 &  1.48 &  0.02 \\
Fear      &  1.31 &  2.44 &  2.99 &  3.53 &  8.40 &  2.42 \\
Sadness   &  2.80 &  4.83 &  6.01 &  6.58 & 12.86 &  4.99 \\
Neutral   &  7.55 & 12.52 & 14.96 & 16.44 & 27.23 & 13.47 \\
Surprise  &  6.37 & 10.62 & 12.68 & 14.04 & 23.74 & 11.12 \\
Happiness & 17.61 & 26.99 & 31.62 & 33.01 & 34.19 & 29.96 \\
\bottomrule
\end{tabular}
}
\caption{Deal rate (\%) by buyer--seller emotion pair, averaged across the five
models and both budget scenarios. Buyer-side emotion strongly affects agreement
formation: angry buyers almost never reach agreement, while happy buyers produce
the highest deal rates.}
\label{tab:agreement-matrix}
\end{table}

\section{Deal Rate by Buyer--Seller Emotion Pair - Emotion Only}

\begin{table}[H]
\centering
\small
\setlength{\tabcolsep}{4pt}
\resizebox{\linewidth}{!}{%
\begin{tabular}{lrrrrrr}
\toprule
\textbf{Buyer $\downarrow$ / Seller $\rightarrow$} & \textbf{Anger} & \textbf{Fear} & \textbf{Sadness} & \textbf{Neutral} & \textbf{Surprise} & \textbf{Happiness} \\
\midrule
Anger     &  0.04 &  0.14 &  0.19 &  0.39 &  1.66 &  0.03 \\
Fear      &  1.46 &  2.08 &  2.55 &  4.04 &  9.34 &  2.06 \\
Sadness   &  2.50 &  4.25 &  5.32 &  7.49 & 11.50 &  5.20 \\
Neutral   &  8.55 & 11.08 & 16.68 & 18.70 & 24.20 & 12.03 \\
Surprise  &  5.56 &  9.51 & 11.27 & 12.40 & 20.65 & 10.78 \\
Happiness & 19.89 & 23.90 & 27.23 & 36.98 & 38.75 & 26.48 \\
\bottomrule
\end{tabular}}
\caption{Deal rate (\%) by buyer--seller emotion pair, averaged across the five
models and both budget scenarios. Buyer-side emotion strongly affects agreement
formation: angry buyers almost never reach agreement, while happy buyers produce
the highest deal rates.}
\label{tab:deal-rate-pairs}
\end{table}
 
% ============================================================
% APPENDIX C  -- cross-model rates (was: RQ4 figure)
% ============================================================
\section{Cross-Model Outcome Rates}
\label{app:rq4-cross-model}
 
\begin{table}[H]
\centering
\small
\setlength{\tabcolsep}{4pt}
\resizebox{\linewidth}{!}{%
\begin{tabular}{lrrrrrr}
\toprule
\textbf{Model} & \textbf{DR} & \textbf{PRR\textsubscript{B}} & \textbf{PRR\textsubscript{S}} & \textbf{Len.} & \textbf{CS} & \textbf{NCS} \\
\midrule
\multicolumn{7}{l}{\textit{Averaged over buyer emotions}} \\
GPT-3.5-Turbo      &  8.35 &  6.21 &  8.60 & 11.33 &   58k & 0.039 \\
GPT-4o-mini        & 12.53 &  9.14 &  4.25 & 10.59 &  129k & 0.086 \\
Gemini 2.5 Flash   & 14.37 &  9.02 &  9.25 & 10.19 &  143k & 0.096 \\
Claude 3.5 Sonnet  & 12.12 &  7.45 &  8.07 & 11.12 &  100k & 0.067 \\
DeepSeek-R1        &  8.98 & 13.31 &  9.38 & 13.95 &  115k & 0.077 \\
\midrule
\multicolumn{7}{l}{\textit{Averaged over seller emotions}} \\
GPT-3.5-Turbo      &  8.86 &  8.69 & 15.76 & 11.57 &   57k & 0.038 \\
GPT-4o-mini        & 13.28 & 12.84 &  8.02 & 10.68 &  124k & 0.083 \\
Gemini 2.5 Flash   & 14.96 & 12.72 & 17.14 & 10.19 &  138k & 0.092 \\
Claude 3.5 Sonnet  & 12.99 & 10.46 & 14.85 & 10.94 &   97k & 0.065 \\
DeepSeek-R1        &  9.56 & 18.59 & 17.25 & 13.95 &  111k & 0.074 \\
\bottomrule
\end{tabular}
}
\caption{Cross-model outcome rates. Rate metrics are percentages; NCS is
$\text{CS}/p_r$. Retained models differ in DRs and negotiation length,
but the qualitative emotion effects reported in RQ1--RQ3 hold across all of them.}
\label{tab:cross-model}
\end{table}
 
% ============================================================
% APPENDIX D  -- negotiation length (was: utterance-length figure)
% ============================================================
\section{Negotiation Length by Model and Buyer Emotion}
\label{app:rq2-length}
 
\begin{table}[H]
\centering
\small
\setlength{\tabcolsep}{4pt}
\resizebox{\linewidth}{!}{%
\begin{tabular}{lrrrrrr}
\toprule
\textbf{Buyer emotion} & \textbf{GPT-3.5-Turbo} & \textbf{GPT-4o-mini} & \textbf{Gemini 2.5} & \textbf{Claude 3.5} & \textbf{DeepSeek-R1} & \textbf{Mean} \\
\midrule
Anger     &  7.47 &  4.12 &  3.31 &  5.33 & 10.17 &  6.08 \\
Fear      & 12.44 & 15.36 & 15.29 & 14.01 & 12.22 & 13.86 \\
Sadness   &  9.58 & 12.18 & 12.37 & 11.02 & 14.00 & 11.83 \\
Neutral   & 11.26 &  9.13 &  9.14 & 10.30 & 13.22 & 10.61 \\
Surprise  & 13.33 & 11.03 & 10.23 & 12.12 & 16.59 & 12.66 \\
Happiness & 13.88 & 11.73 & 10.82 & 13.96 & 17.48 & 13.57 \\
\bottomrule
\end{tabular}
}
\caption{Average negotiation length (turns) by model and buyer emotion. Emotion
conditioning changes the length of the interaction: angry buyers terminate quickly,
while fearful and happy buyers sustain markedly longer negotiations.}
\label{tab:rq2-length}
\end{table}
 
% ============================================================
% APPENDIX E  -- termination profile (was: figure)
% ============================================================
\section{Termination Profile by Buyer Emotion}
\label{app:rq2-termination}
 
\begin{table}[H]
\centering
\small
\begin{tabular}{lrrrr}
\toprule
\textbf{Buyer emotion} & \textbf{Accepted} & \textbf{Rejected} & \textbf{Deadlock} & \textbf{Total} \\
\midrule
Anger     &  0.39 & 87.19 & 12.42 & 100.00 \\
Fear      &  3.52 & 54.91 & 41.57 & 100.00 \\
Sadness   &  6.35 & 59.12 & 34.53 & 100.00 \\
Neutral   & 15.33 & 56.94 & 27.73 & 100.00 \\
Surprise  & 13.09 & 55.73 & 31.18 & 100.00 \\
Happiness & 28.91 & 30.27 & 40.82 & 100.00 \\
\bottomrule
\end{tabular}
\caption{Termination profile (\%) by buyer emotion, averaged across the five models,
all seller emotions, and both budget scenarios. Angry buyers overwhelmingly reject,
while happy buyers produce the highest DR and fearful buyers the highest
deadlock rate.}
\label{tab:rq2-termination}
\end{table}
 
\clearpage
\onecolumn
 
% ============================================================
% APPENDIX F  -- full buyer-side results
% ============================================================
\section{Complete Results by Model and Buyer Emotion}
\label{app:buyer-results-tables}
 
\begin{table}[H]
\centering
\caption{Results by model and buyer emotion, averaged across seller emotions and
budget scenarios. DR, PRR\textsubscript{B} and PRR\textsubscript{S} are percentages;
Len.\ is in turns; Profit and CS are compactly formatted (e.g.\ \texttt{30k}).}
\label{tab:appendix-buyer}
\setlength{\tabcolsep}{5pt}
\small
\begin{tabular}{llrrrrrr}
\toprule
\textbf{Model} & \textbf{Buyer Emo.}
  & \textbf{DR} & \textbf{PRR\textsubscript{B}} & \textbf{PRR\textsubscript{S}}
  & \textbf{Profit} & \textbf{Len.} & \textbf{CS} \\
\midrule
GPT-3.5-Turbo      & Anger     &   0.88 &   0.01 &    4.20 &       1.5k &  7.47 &   91k \\
GPT-4o-mini        & Anger     &   0.01 &   0.00 &    0.00 &          0 &  4.12 &  238k \\
Gemini 2.5 Flash   & Anger     &   0.22 &   0.00 &    1.80 &       1.2k &  3.31 &  270k \\
Claude 3.5 Sonnet  & Anger     &   0.21 &   0.02 &    2.80 &        787 &  5.33 &  171k \\
DeepSeek-R1        & Anger     &   0.63 &   0.06 &    2.20 &       1.0k & 10.17 &  210k \\
\midrule
GPT-3.5-Turbo      & Fear      &   3.43 &  12.82 &    7.30 &      28.8k & 12.44 &   54k \\
GPT-4o-mini        & Fear      &   4.90 &  17.04 & $-$4.10 &  $-$711.7k & 15.36 &   97k \\
Gemini 2.5 Flash   & Fear      &   5.58 &  17.18 &    5.50 &      22.8k & 15.29 &  118k \\
Claude 3.5 Sonnet  & Fear      &   3.58 &  16.32 &    5.70 &       9.4k & 14.01 &   82k \\
DeepSeek-R1        & Fear      &   0.11 &  23.68 &    3.60 &      16.1k & 12.22 &   90k \\
\midrule
GPT-3.5-Turbo      & Sadness   &   1.24 &   3.28 &    2.40 &         13 &  9.58 &   86k \\
GPT-4o-mini        & Sadness   &   9.76 &  13.91 &    1.30 &  $-$375.3k & 12.18 &  136k \\
Gemini 2.5 Flash   & Sadness   &  12.21 &  14.66 &   10.40 &      40.6k & 12.37 &  153k \\
Claude 3.5 Sonnet  & Sadness   &   6.73 &   7.32 &    4.20 &      14.7k & 11.02 &  117k \\
DeepSeek-R1        & Sadness   &   1.79 &  10.86 &    4.60 &       7.4k & 14.00 &  130k \\
\midrule
GPT-3.5-Turbo      & Neutral   &   9.84 &   2.81 &    9.20 &    $-$4.0k & 11.26 &   33k \\
GPT-4o-mini        & Neutral   &  19.45 &  12.67 &    2.70 &  $-$326.1k &  9.13 &  120k \\
Gemini 2.5 Flash   & Neutral   &  22.82 &  10.75 &    8.90 &      96.5k &  9.14 &  140k \\
Claude 3.5 Sonnet  & Neutral   &  16.88 &   7.52 &    6.50 &      53.6k & 10.30 &   79k \\
DeepSeek-R1        & Neutral   &   7.74 &   9.37 &   12.20 &      24.8k & 13.22 &  100k \\
\midrule
GPT-3.5-Turbo      & Surprise  &  10.59 &   9.12 &    8.10 &      35.3k & 13.33 &   45k \\
GPT-4o-mini        & Surprise  &  13.97 &   1.38 &   10.70 &     196.7k & 11.03 &  104k \\
Gemini 2.5 Flash   & Surprise  &  17.18 &   1.36 &   12.20 &     293.9k & 10.23 &   96k \\
Claude 3.5 Sonnet  & Surprise  &  14.57 &   4.03 &   10.10 &     135.6k & 12.12 &   87k \\
DeepSeek-R1        & Surprise  &   9.12 &  24.11 &   10.30 &     166.2k & 16.59 &   90k \\
\midrule
GPT-3.5-Turbo      & Happiness &  24.10 &   9.24 &   20.40 &     275.4k & 13.88 &   42k \\
GPT-4o-mini        & Happiness &  27.11 &   9.85 &   14.90 &     118.6k & 11.73 &   77k \\
Gemini 2.5 Flash   & Happiness &  28.18 &  10.16 &   16.70 &     169.7k & 10.82 &   83k \\
Claude 3.5 Sonnet  & Happiness &  30.72 &   9.51 &   19.10 &     172.3k & 13.96 &   67k \\
DeepSeek-R1        & Happiness &  34.46 &  11.79 &   23.40 &     171.0k & 17.48 &   70k \\
\bottomrule
\end{tabular}
\end{table}

\section{Human Judgment}

\begin{table}[ht]
\centering
\small
\setlength{\tabcolsep}{4pt}
\begin{tabular}{lcccccc}
\toprule
Model & Anger & Fear & Sadness & Neutral & Surprise & Happiness \\
\midrule
GPT-3.5-Turbo     & 76.48 & 71.88 & 72.21 & 72.90 & 73.68 & 74.63 \\
GPT-4o-mini       & 84.22 & 78.91 & 81.18 & 83.83 & 79.65 & 84.73 \\
Gemini 2.5 Flash  & 82.49 & 80.17 & 77.76 & 85.93 & 75.91 & 81.58 \\
DeepSeek-R1       & 77.81 & 74.51 & 76.37 & 81.70 & 79.53 & 76.94 \\
Claude Sonnet 3.52 & 83.64 & 75.65 & 81.34 & 85.38 & 81.43 & 86.30 \\
\bottomrule
\end{tabular}
\caption{Agreement (\%) between the automated Judge and human annotations of
the negotiation state $d_t \in \{\textsc{accept}, \textsc{reject},
\textsc{continue}\}$.}
\label{tab:judge-agreement}
\end{table}
\newpage
\section{Hyper-parameters}
 \begin{table}[ht]
\centering
\small
\setlength{\tabcolsep}{4pt}
\begin{tabular}{llcc}
\toprule
Model & Access  & top-$p$ & top-$k$ \\
\midrule
GPT-3.5-Turbo     & OpenAI API    & 1.0   & --    \\
GPT-4o-mini       & OpenAI API     & 1.0   & --    \\
Gemini 2.5 Flash  & Google API     & 0.95  & 64    \\
Claude 3.5 Sonnet & Anthropic API  & unset & unset \\
DeepSeek-R1       & Ollama         & 0.9   & 40    \\
\midrule
\multicolumn{4}{l}{\emph{Auxiliary (analyst, judge)}} \\
GPT-4o-mini       & OpenAI API     & 1.0   & --    \\
\bottomrule
\end{tabular}
\caption{All values are provider defaults except temperature, which is set to $\tau=0$ for determinism.}
\label{tab:hyperparams}
\end{table}

\end{document}